\documentclass[letterpaper, 10 pt, journal, twoside]{IEEEtran}
\ifCLASSINFOpdf
\else
\fi
\usepackage{times}
\usepackage{multicol}
\usepackage{hyperref}
\usepackage{graphicx}
\usepackage{bm}
\usepackage[nolist]{acronym}
\usepackage{comment}
\usepackage{amsmath}
\usepackage{amssymb}
\usepackage[capitalise]{cleveref}
\usepackage{algorithm}
\usepackage[usenames,dvipsnames]{xcolor}
\usepackage{algorithmic}
\usepackage{booktabs}
\usepackage{array}
\usepackage{dblfloatfix}
\usepackage{tabularx}
\usepackage{enumitem}
\usepackage{subcaption}
\usepackage{diagbox}
\usepackage{etoolbox}\AtBeginEnvironment{algorithmic}{\small}

\usepackage{fancyhdr}
\begin{document}

\title{Aggressive Perception-Aware Navigation using \\ Deep Optical Flow Dynamics and PixelMPC}

\author{Keuntaek~Lee,~Jason~Gibson, and~Evangelos~A.~Theodorou%
\thanks{Manuscript received: September, 10, 2019; Revised December, 6, 2019; Accepted January, 6, 2020.}%Use only for final RAL version
\thanks{This paper was recommended for publication by Editor Eric Marchand upon evaluation of the Associate Editor and Reviewers' comments.
This work was supported by NASA.} %Use only for final RAL version
\thanks{The authors are with the Autonomous Control and Decision Systems Laboratory,
Georgia Institute of Technology, Atlanta, GA 30332, USA.
        {\tt\footnotesize keuntaek.lee@gatech.edu}}%
}

% \markboth{IEEE Robotics and Automation Letters. Preprint Version. January, 2020}
% {Lee \MakeLowercase{\textit{et al.}}: Aggressive Perception-Aware Navigation using Deep Optical Flow Dynamics and PixelMPC} 

\begin{acronym}
\acro{NN}{Neural Network}
\acro{CNN}{Convolutional Neural Network}
\acro{DL}{Deep Learning}
\acro{IL}{Imitation Learning}
\acro{iLQG/MPC-DDP}{iterative Linear Quadratic Gaussian/Model Predictive Control Differential Dynamic Programming}
\acro{MPC}{Model Predictive Control}
\acro{DDP}{Differential Dynamic Programming}
\acro{BNN}{Bayesian Neural Network}
\acro{PixelMPC}{Pixel Model Predictive Control}
\end{acronym}

\maketitle
% for headers and footers on the initial page
\thispagestyle{fancy}
\pagestyle{empty}

\begin{abstract}
Recently, vision-based control has gained traction by leveraging the power of machine learning. In this work, we couple a model predictive control (MPC) framework to a visual pipeline. We introduce deep optical flow (DOF) dynamics, which is a combination of optical flow and robot dynamics. Using the DOF dynamics, MPC explicitly incorporates the predicted movement of relevant pixels into the planned trajectory of a robot. Our implementation of DOF is memory-efficient, data-efficient, and computationally cheap so that it can be computed in real-time for use in an MPC framework. The suggested Pixel Model Predictive Control (PixelMPC) algorithm controls the robot to accomplish a high-speed racing task while maintaining visibility of the important features (gates). This improves the reliability of vision-based estimators for localization and can eventually lead to safe autonomous flight. The proposed algorithm is tested in a photorealistic simulation with a high-speed drone racing task.
\\ Supplementary video: \textcolor{blue}{\url{https://youtu.be/NzL2YRcOh_I}}
\end{abstract}

\begin{IEEEkeywords}
Model Learning for Control, Optimization and Optimal Control, Visual Servoing, Visual Tracking, Visual-Based Navigation
\end{IEEEkeywords}

\IEEEpeerreviewmaketitle

\section{Introduction}
\IEEEPARstart{W}{e} introduce a novel mechanism which combines vision into a model predictive control (MPC) framework.

% Deep learning (DL)-based perceptual control using end-to-end imitation learning has shown great success in many robotics disciplines including autonomous driving \cite{PanRSS18, bojarski2016end, zhang17aaai}, manipulation \cite{LevineVisuomotor}, and autonomous drone flying \cite{giusti2016machine, Smolyanskiy2017TowardLA}. 
Deep learning (DL)-based perceptual control using end-to-end imitation learning has shown great success in many robotics disciplines including autonomous driving \cite{PanRSS18, bojarski2016end, zhang17aaai}, manipulation \cite{LevineVisuomotor}, and autonomous drone flying \cite{giusti2016machine, Smolyanskiy2017TowardLA}.

In this paper, instead of taking a fully end-to-end approach (\cite{giusti2016machine, Smolyanskiy2017TowardLA}), we deploy the power of DL in novel system modeling.
In a traditional (not end-to-end) navigation, DL-aided vision pipeline played a big role in detecting objects and obstacles as a perception module and sometimes as a part of state estimation (e.g. VSLAM \cite{murORB2}). A controller then performed its task of navigation, avoidance, or tracking using the information provided from the vision part \textcolor{black}{\cite{Kaufmann19Beauty}}.

The visual object tracking or visual servoing technologies have been developed over the past few decades and can be found in some commercial drone products.
However, most of the work in literature \cite{giusti2016machine, Smolyanskiy2017TowardLA, Kaufmann18DeepDrone} are all based on reactive controllers; the robot turns left if the object is on the right-side of a robot's view, and vice versa. This reactive visual servoing requires the drone to fly at a slow speed or hover until it finishes servoing.
Here we propose a predictive visual tracking controller for high-speed racing with a data-driven optical flow dynamics model composed of optical flow and robot dynamics.

In a drone racing scenario, the optical flow mostly comes from a moving camera and a static environment. Since the controller moves the robot through space, the changes in scene, the optical flow, can be thought of as \textcolor{black}{indirect} dynamics.
%Since the change of a scene and the optical flow is coming from the quadrotor's movement, which is coming from a controller, the optical flow is in fact coming from control actions. 
%Not only the control but the orientation of a robot affects the optical flow when it moves, so the optical flow dynamics is highly related to the robot dynamics as well.

\begin{figure}[t]
  \centering
  \includegraphics[width=0.3\textwidth]{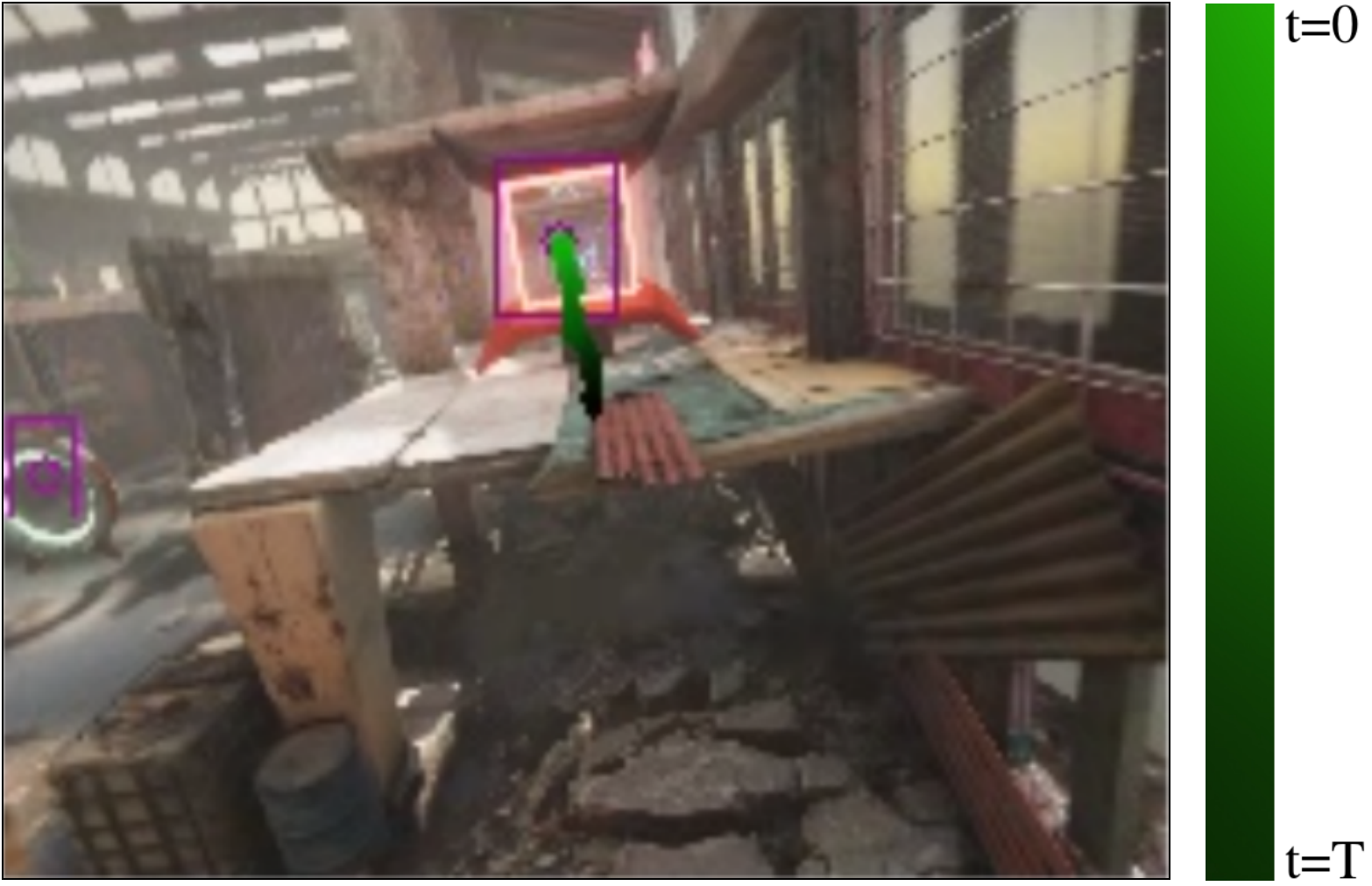}
  \caption{MPC-predicted future pixel trajectory (Green) of a target pixel, the center of a gate. PixelMPC computes the optimal control which accomplishes a racing task and drives the target pixel to the center of the image.}
  \label{fig:pixeltraj}
   \vspace{-0.3cm}
\end{figure}

Recently, there has been a lot of progress in DL-based optical flow techniques \cite{flownet2, Sun_2018_CVPR, Ranjan2017SpyNet}. However, all prior work relies on large convolutional neural networks with a lot of parameters to estimate the optical flow of the entire image. In our work, the application of the optical flow is to \textcolor{black}{predict the relative motion of} a `single' pixel, so we use a small fully-connected feedforward network. 
%It is not right to compare against the deep-learning based optical flow prediction work in the literature because they predict the whole image of the optical flow, but our proposed network fits perfectly in our problem statement with great accuracy of predicting the optical flow.
%Our architecture is fundamentally different and should not be compared to prior results. We will discuss about more details in \cref{sec:DeepOpticalFlowDynamics}.

The main problem we address in this paper is the visibility/field of view of a moving camera, especially when it comes to high-speed racing. The more the robot observes through a camera, the more information we use to perform accurate state estimation and navigation. Therefore, it is important to control the robot to see more information, for example, by pitching up or rolling/yawing. However, this conflicts with the high-speed flying task for a drone because a quadrotor needs to pitch down to fly at a high speed and this results in losing more visual information.
% Similarly, there are other scenarios where modification of other control inputs would result in similar result.
% Also, gaining more visibility of important features cannot be accomplished just by pitching up/down.
%Besides, having more visibility is not a simple pitching up/down problem: there are a lot of other scenarios where yaw or roll angles affects the visibility of important features. For example, if a drone has to make a quick u-turn to fly through gates, the more rapid yaw angular velocity will be required to see more gates in its view.

To solve the problem of limitation in the field of view by visual servoing, \cite{Penin2017MinTime, Murali2019DiffFlat} proposed a Sequential Quadratic Programming-based approach where the visibility is formulated in hard constraints. However, these methods do not fit into our problem formulation which requires real-time planning and control.

In the visual servoing literature, to the best of our knowledge, the real-time predictive controllers used for a visual tracking task are \cite{Negeli2017ViewpointOpt, falanga2018pampc}. Although \cite{Negeli2017ViewpointOpt} formulated an MPC problem for a viewpoint optimization, the goal of the paper was controlling the drone to stabilize a gimbal to get a good quality of a video. In \cite{falanga2018pampc}, the most relevant work to us, the authors derived the target pixel velocity based on the information of the relative 3D position ($x, y, z$) of the target and the robot. With the pixel velocity information, the authors were able to form an MPC problem along with vision and perform visual object tracking control in a predictive way.

However, in our work, we implement a data-driven deep-learning approach, that does not require any prior information of camera intrinsics, extrinsics, or the 3D global position of the target. Instead, our algorithm requires an object detector that detects a target in image space. Thanks to the great success in the field of computer vision, we can use real-time object detectors \cite{yolov3,fasterrcnn} with GPUs. Although our method requires prior knowledge of \textcolor{black}{the image of} the targets (gates) and a trained detector, we believe this is less restrictive than full knowledge of the global 3D position of features like in \cite{falanga2018pampc}. Furthermore, we believe our case is less restrictive since our proposed approach can be used for any \textcolor{black}{moving} target objects located anywhere in the scene.
%Our approach is, therefore, more adaptive and flexible compared to having prior knowledge of the world-frame position of the target objects.

% If we want to mix two different controllers for two different tasks, it is hard to optimize between them. However, our approach fuses two different dynamics into one optimization formulation and derive a single optimal controller which accomplishes two different tasks in two different spaces (3D world space and 2D image space).

In summary, the contributions of this work are twofold: 
\begin{itemize}[leftmargin=0.4cm]
  \item We introduce data-driven Deep Optical Flow (DOF) dynamics, learned from the optical flow of consecutive images and robot dynamics. DOF dynamics are efficient in memory and computation.
  \item We introduce the Pixel Model Predictive Control (PixelMPC) algorithm which \textcolor{black}{predicts the relative motion of pixels} by actuating the robot to visually track important features (targets) while accomplishing the high-level tasks (e.g. racing or chasing). The algorithm makes the vision-based state estimation more robust as it 
  %allows the robot to have more important visual information.
  explicitly allows the control algorithm to prioritize visual information.
\end{itemize}

The remaining of the paper is organized as follows: In \cref{sec:Preliminaries}, we briefly review some preliminaries used in our work. The DOF dynamics are introduced in \cref{sec:DeepOpticalFlowDynamics} and in \cref{sec:PixelMPC}, we introduce our PixelMPC algorithm. \cref{sec:Experiments} details vision-based drone racing and state estimation experiments with analysis and comparisons of the proposed methods. Finally, we conclude and discuss future directions in \cref{sec:conclusion}.

\section{Preliminaries}
\label{sec:Preliminaries}

In this section, we provide the building blocks of the proposed \ac{PixelMPC}: MPC, drone dynamics model, and the optical flow.

\subsection{Model Predictive Optimal Control}
\label{subsec:MPC}
\ac{MPC}-based optimal controllers (e.g. Model Predictive Path Integral (MPPI) \cite{mppi}) provide planned control trajectories given an initial state and a cost function by solving the optimal control problem. An optimal control problem whose objective is to minimize a task-specific cost function $J(\mathbf{X}, \mathbf{U})$ can be formulated as follows:
\begin{align}
    J(\mathbf{X}(t), \mathbf{U}(t)) &= \phi(\mathbf{X}(t_f)) + \int_{t=t_0}^{t_f}l(\mathbf{X}(t), \mathbf{U}(t))dt \label{eq:costfunction} \\ 
	V(\mathbf{X}(t_0), t_0) &= \min_{\mathbf{U}(t)}\Big[ J(\mathbf{X}(t), \mathbf{U}(t)) \Big] \label{eq:optcontrolprob}
\end{align}
subject to dynamics
\begin{align}
    \frac{d\mathbf{X}}{dt} = F(\mathbf{X}(t), \mathbf{U}(t), t),  
\end{align}
where $\mathbf{X} \in \mathbb{R}^n$ represents the system states, $\mathbf{U} \in \mathbb{R}^m$ represents the control, $\phi$ is the state cost at the final time $t_f$, $l$ is the running cost, and $V$ is the value function. 
By solving this local optimization problem, we get the optimal control sequences. This can be solved in a receding horizon fashion in an MPC framework and it allows us to have a real-time optimal controller with feedback.

In our work, a sampling-based receding-horizon stochastic optimization algorithm, MPPI controller \cite{mppi} is used as an MPC controller. We chose MPPI for several reasons, first off being the generality of cost functions and dynamics allowed. Most variants of MPC require us to have a convex cost function and first or second-order approximations of the dynamics. MPPI has neither of these requirements. Therefore we can directly encode our task into the cost function without any modifications to the high-level objective. Second, MPPI has been shown to be highly successful at aggressive autonomous racing on ground vehicles with general cost functions and neural network dynamics \cite{mppi}.

For a short summary of MPPI algorithmically, it samples $N$ trajectories by applying noise into the control channels and forward propagating the dynamics.
% The trajectories are parameterized by time, where each step is $\Delta t$, for a total of $T$ steps. The control vector noise is parameterized by a zero mean Gaussian distribution ($0, \sigma$), where each $\sigma$ can be independently varied.
Each sample can be rolled out in parallel, and then each corresponding trajectory and cost are combined to generate a final control vector. The optimization can be run $K$ times to further refine the solution before executing it.
The previous control solution is used as the center value of the Gaussian sampling to warm start the optimization each round.

% Our work with this algorithm will be entirely on the level of cost function design and novel dynamics. %The underlying optimization done by MPPI will remain the same as in \cite{mppi}.

%Out of several MPC algorithms, MPPI was used in our work because it has a great ability to optimize non-quadratic cost functions, which fits well in our case where we design a simple pixel position-dependent cost function (\cref{eq:pixelcost}).

\subsection{Quadrotor Dynamics}
\label{sec:dronedynamics}
\begin{figure}[ht]
    \centering
    \includegraphics[width=0.4\textwidth]{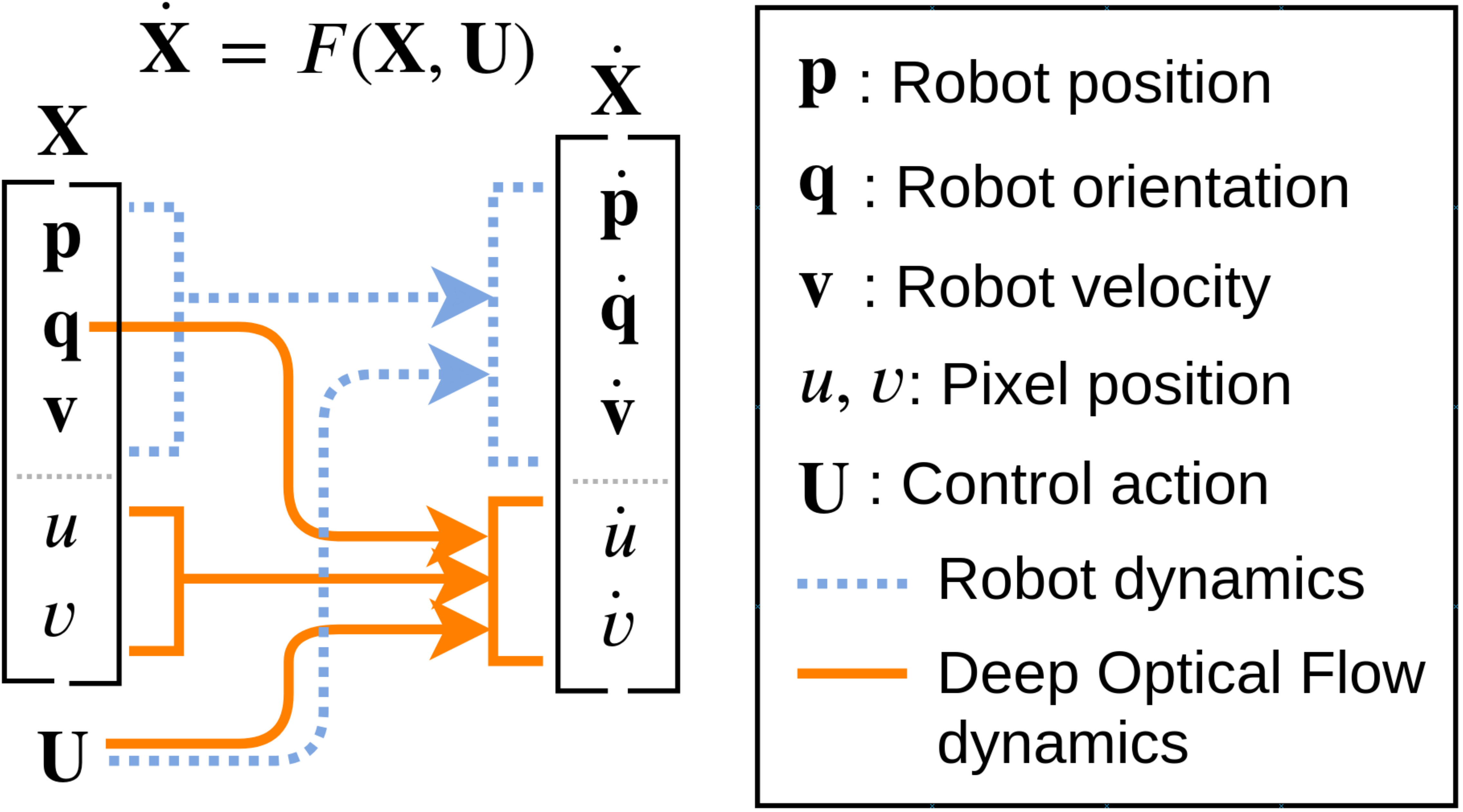}
    \caption{The total model dynamics $\dot{\mathbf{X}}=F(\mathbf{X}, \mathbf{U})$ used in the PixelMPC. The model is composed of the deep optical flow (DOF) dynamics and robot dynamics.}\label{fig:fx}
\end{figure}

We use the quadrotor dynamics model provided in the FlightGoggles simulator \cite{FlightGoggles} used in this paper. The defined 10 states are $\mathbf{X}_{\textcolor{black}{\text{robot}}}=[\mathbf{p}; \mathbf{q}; \mathbf{v}]=[x, y, z, q_w, q_x, q_y, q_z, \dot{x}, \dot{y}, \dot{z}]^T$, where $\mathbf{p}=[x, y, z]^T$ is the world-coordinate position vector, $\mathbf{q}=[q_w, q_x, q_y, q_z]^T$ is the vehicle attitude unit quaternion vector, and $\mathbf{v}=\dot{\mathbf{p}}=[\dot{x}, \dot{y}, \dot{z}]^T$ is the world-coordinate linear velocity vector. The vehicle dynamics are given by
\begin{align}
    \dot{\mathbf{p}} &= \mathbf{v} \label{eq:dronedynamics1} \\
    \dot{\mathbf{v}} &= \mathbf{g} + m^{-1}(\mathbf{R}_b^\omega \mathbf{f}_T \textcolor{black}{ + } \mathbf{f}_D + \mathbf{w_f}) \label{eq:dronedynamics2},
\end{align}
where $\mathbf{g}$ is the gravitational acceleration, $m$ is the quadrotor mass, $\mathbf{R}_b^\omega$ is the rotation matrix from body to world frame, $\mathbf{f}_T$ is the total thrust, $\mathbf{f}_D$ is the aerodynamic drag, and $\mathbf{w_f}$ is the stochastic force vector to capture unmodeled dynamics (e.g. vibrations and turbulance). The rotation matrix from body to world frame is
\begin{align}
    \mathbf{R}_b^\omega =
    \begin{bmatrix}
    1-2(\textcolor{black}{q_y}^2+q_z^2) & 2(q_x q_y-q_z q_w) & 2(q_x q_z + q_y \textcolor{black}{q_w}) \\
    2(q_x q_y + q_z q_w) & 1-2(q_x^2+q_z^2) & 2(q_y q_z - q_x q_w) \\
    2(q_x q_z - q_y q_w) & 2(q_y q_z + q_x q_w) & 1-2(q_x^2+q_y^2)
    \end{bmatrix}, \label{eq:dronedynamics3}
\end{align}
and the \textcolor{black}{relation between quaternions and the angular rates} is
\begin{align}
    \dot{\mathbf{q}} = \frac{1}{2}\begin{bmatrix}
    -q_x & -q_y & -q_z \\
    q_w & -q_z & q_y \\
    \textcolor{black}{q_z} & q_w & -q_x \\
    -q_y & q_x & q_w
    \end{bmatrix}
    \begin{bmatrix}
    \omega_x \\ \omega_y \\ \omega_z
    \end{bmatrix}, \label{eq:dronedynamics4}
\end{align}
where the angular rates $\omega_x, \omega_y,$ and $\omega_z$ are \textcolor{black}{part of} the control inputs we used along with the total thrust $\mathbf{f}_T$. \textcolor{black}{The control $\mathbf{U}$ is $[\omega_x, \omega_y, \omega_z, \mathbf{f}_T]$}. We make a small assumption here that the model immediately follows the control inputs, especially the angular rates. Indeed, the quadrotor in the FlightGoggles takes $\mathbf{U}$ as an input and the low-level PID controller controls the robot to follow the commands. Since we directly input the angular rates, we do not use the dynamics of the angular rates, described in \cite{FlightGoggles} when we propagate the model in MPC. The robot dynamics we used in this paper is also described in \cref{fig:fx}.

\subsection{Optical Flow}
\label{subsec:OpticalFlow}
\textcolor{black}{Optical flow estimates the instantaneous motion of objects and features in a visual scene from a sequence of ordered images.} The motion comes from the relative motion between an observer and a scene. In our case, the motion comes from a moving observer (a camera attached on a robot) and a static environment.
To compute the optical flow, two strict assumptions are required: 1) The brightness of any observed object point on images is constant over time, 2) In the image plane, neighborhood points move similarly with similar velocity. The first constraint can be written as:
\begin{align}
    I(u,v,t) = I(u+\Delta u, v+\Delta v, t+\Delta t),
\end{align}
where $I$ represents the intensity of a pixel $(u, v)$ and $\Delta u, \Delta v$ represent the displacement of the pixel position between two consecutive images observed at time $t$ and $t+\Delta t$.
This equation can be written in a form of Taylor series by assuming that the movement is small:
\begin{align}
    I(u+\Delta u, v+\Delta v, t+\Delta t) &= I(u, v, t) + \frac{\partial I}{\partial u} \Delta u + \frac{\partial I}{\partial v} \Delta v \nonumber \\
    &\hspace{0.5cm}+ \frac{\partial I}{\partial t} \Delta t + H.O.T,
\end{align}
which results in 
\begin{align}
    \frac{\partial I}{\partial u} \frac{\Delta u}{\Delta t} + \frac{\partial I}{\partial v} \frac{\Delta v}{\Delta t} + \frac{\partial I}{\partial t} = 0.
\end{align}

However, it is impossible to estimate the two unknowns $\frac{\Delta u}{\Delta t}$ and $\frac{\Delta v}{\Delta t}$, only with one equation, so all the optical flow calculation methods make additional assumptions to estimate the actual flow.

\begin{figure}
\begin{subfigure}{.15\textwidth}
  \centering
  \includegraphics[width=.6\textwidth]{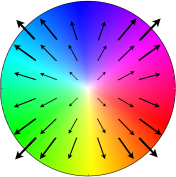}
  \caption{Legend of optical flow}
  \label{fig:optical_flow_legend}
\end{subfigure}%
\vspace{2mm}
\begin{subfigure}{.15\textwidth}
  \centering
  \includegraphics[width=.9\textwidth]{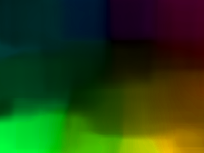}
  \caption{Optical flow ground truth}
  \label{fig:deep_optical_flow_gt}
\end{subfigure}
\vspace{2mm}
\begin{subfigure}{.15\textwidth}
  \centering
  \includegraphics[width=.9\textwidth]{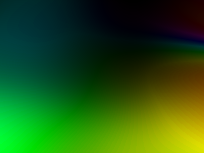}
  \caption{Deep optical flow (DOF) prediction}
  \label{fig:deep_optical_flow_nn}
\end{subfigure}
\caption{Ground truth optical flow and the DOF prediction in the case of a quadrotor flying forward, pitching down. Deep optical flow provides more smooth optical flow compared to the ground truth. DOF predicts a single pixel's optical flow instead of the whole image's flow.}
\label{fig:deep_optical_flow}
\vspace{-0.5cm}
\end{figure}

We used one of the most popular algorithms \cite{Farneback03} to calculate the dense optical flow. The algorithm approximates each neighborhood of both frames by quadratic polynomials. 
% The polynomial approximations were done separately for the first image and the second image. The displacement of the pixel position is also parameterized by a linear model. 
The details of the algorithm can be found in \cite{Farneback03} and the implementation of the algorithm is available in OpenCV \cite{opencv}. 
For a better calculation of the dense optical flow, we used a sequence of downsized gray-scaled images instead of original RGB images.
The parameters used for calculating optical flow with \cite{Farneback03} were: pyr scale=0.5, levels=10, winsize=51, iterations=15, poly$_n$=5, poly$_\sigma$=1.1.
The visualization of the dense optical flow as a vector or in color can be found in \cref{fig:deep_optical_flow} and the supplementary video includes the optical flow of a full run of racing.

\section{Deep Optical Flow Dynamics}
\label{sec:DeepOpticalFlowDynamics}
By taking advantage of the algorithms \cite{Farneback03} calculating the optical flow, deep optical flow learning becomes self-supervised learning, which does not require any manual labeling.
Our proposed neural network-based Deep Optical Flow (DOF) dynamics have two major selling points:
\subsubsection{Computationally efficient}
DOF dynamics predict an optical flow/vector of a single pixel while most of the DL-based optical flow \textcolor{black}{\cite{flownet2, Sun_2018_CVPR, Ranjan2017SpyNet}} predicts the next timestep's image of optical flow, with the same size of the input images. This allows us to have a very small network, so we can use the model in a real-time optimal controller that performs optimization within 20-50ms. If we build a U-Net-like convolutional neural network, which predicts an image from an input image, we have to propagate the deep CNN every timestep in MPC framework to generate optical flow, which is computationally very expensive and slow. For the parameters used in the paper, our MPC algorithm samples over a million times per second. 
%Our method does not rely on convolutions and uses a simple 3-layer feedforward network in order to run in real-time.
\subsubsection{Data-efficient}
Given an image, size of W$\times$H, DOF can use W$\times$H data points for training, whereas typical DL-based optical flow \textcolor{black}{\cite{flownet2, Sun_2018_CVPR, Ranjan2017SpyNet}} only uses a single data point (an image of the whole optical flow).

\begin{figure}
\begin{subfigure}{.13\textwidth}
  \centering
  \includegraphics[width=\textwidth]{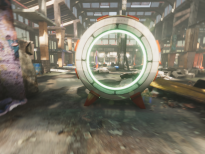}
\end{subfigure}
\begin{subfigure}{.13\textwidth}
  \centering
  \includegraphics[width=\textwidth]{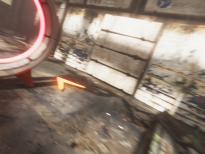}
\end{subfigure}
\begin{subfigure}{.13\textwidth}
  \centering
  \includegraphics[width=\textwidth]{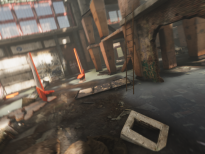}
\end{subfigure}
\begin{subfigure}{.13\textwidth}
  \centering
  \includegraphics[width=\textwidth]{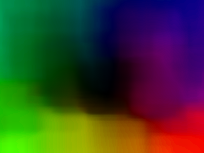}
  \caption{Fly forward}
\end{subfigure}
~~~~~~
\begin{subfigure}{.13\textwidth}
  \centering
  \includegraphics[width=\textwidth]{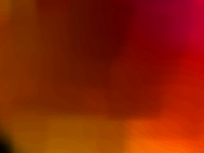}
  \caption{Fly left}
\end{subfigure}
~~~~~~
\begin{subfigure}{.13\textwidth}
  \centering
  \includegraphics[width=\textwidth]{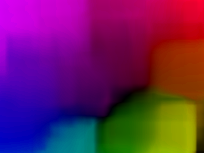}
  \caption{Roll clockwise}
\end{subfigure}
\caption{Optical flow depending on the robot orientation and control. For a legend of the colormap, please refer to \cref{fig:optical_flow_legend}.}
\label{fig:optical_flow_orientation}
\vspace{-0.5cm}
\end{figure}

DOF dynamics predict, just like typical robot dynamics models, the derivative of the states. Here, in DOF, it predicts the velocity of a pixel. DOF takes 3 components as input: pixel state (position) $\mathbf{X}_{\text{pixel}}$, control actions $\mathbf{U}$, and robot orientation $\mathbf{q}$.
The pixel position means the position in (u, v) coordinate system on the image plane, where the top left corner is the origin (0, 0). Control actions command angular velocities in $x,y,z$ frame and total thrust, which affects both robot motion/acceleration and the image stream. The main point here in the DOF input is the robot orientation part. We incorporate the orientation of the robot into the DOF dynamics because even with the same control input, the optical flow changes depending on the roll, pitch, and yaw angles of the robot as shown in \cref{fig:optical_flow_orientation}.

We train the DOF dynamics with a neural network (NN) model to predict the magnitude $l$ and the angle $\theta$ of a single optical flow/vector.
By defining the state of the pixel $\mathbf{X}_{\text{pixel}} = [u, v]^T$, we can write the optical flow as
\begin{align}\label{eq:polar2euler}
    \dot{u} = l cos(\theta) , ~~ \dot{v} = l sin(\theta),
\end{align}
where $l$ and $\theta$ are the optical flow vector component, predicted from the DOF. Therefore, the final DOF dynamics $F_{\text{pixel}}$ is
\begin{align}
    \dot{\mathbf{X}}_{\text{pixel}} &= F_{\text{pixel}}(\mathbf{q}, \mathbf{X}_{\text{pixel}}, \mathbf{U}) \\
    &= \text{PolarToEuler}(DOF(\mathbf{q}, \mathbf{X}_{\text{pixel}}, \mathbf{U})), \label{eq:DOFdynamics}
\end{align}
where the PolarToEuler mapping is \cref{eq:polar2euler}. % The proposed dynamics can be also found in \cref{fig:fx}.

\cref{alg:trainingdeepopticalflow} describes the training process of DOF dynamics.
\textcolor{black}{In the first for-loop of \cref{alg:trainingdeepopticalflow}, the robot state $\mathbf{X}_{\text{robot}}$ can be either ground truth or estimated states. The first for-loop describes collecting training dataset of robot states, optimal control actions, images, and optical flow between two consecutive images. Then the following for-loops update the weights and biases of DOF dynamics model with respect to the mean squared error (MSE) loss between the target magnitude and the angle of optical flow and the prediction.}

We normalize the pixel state into [0.0, 1.0]$\times$[0.0, 1.0] space and do regression. This allows the original discrete image space [0, W]$\times$[0, H] to be a continuous 2D space [0.0, 1.0]$\times$[0.0, 1.0] and same for the pixel state space, as well.

We designed a feed-forward NN with 5 layers having [10, 128, 128, 128, 2] neurons per each, where 10 is for an input layer and 2 is for the output. The \textcolor{black}{Rectified Linear Unit ($ReLU$) function, $f(x)=max(0, x)$,} is used for the activation function in layers 1-4, and the output layer has a linear activation.
\textcolor{black}{All the layers are fully connected with regularization via 10$\%$ of dropouts. The motivation behind using the stated number of neurons was to achieve real-time performance with MPPI.
We had to balance the accuracy of the model with real-time constraints. A more accurate model would need more neurons and more layers, but this would prevent real-time usage in MPPI. We empirically choose an architecture that was accurate and could still be run in real-time.
These two goals conflicts: To achieve a more accurate model, we need more neurons and more layers, but this would result in slow inference time. As a result, we empirically chose the numbers to achieve both goals.
}
For training the neural network, the Adam \cite{adam} optimizer was used with Tensorflow \cite{tensorflow}.

The usage of the trained model can be found in \cref{alg:testingdeepopticalflow}. \textcolor{black}{Given a center of the object $(u, v)$ from a Detector (e.g. YOLOv3 \cite{yolov3}), a trained DOF dynamics model takes the center position ($u, v$), robot orientation, and control action as an input. The output of the trained DOF dynamics is the magnitude $l$ and the angle $\theta$ of a predicted optical flow of that single point $(u, v)$. From predicted $l$ and $\theta$, the velocity of the single point is calculated as in \cref{eq:polar2euler}.}

\begin{algorithm} [ht]
\caption{Training Deep Optical Flow (DOF) Dynamics}
\begin{algorithmic}[1] \label{alg:trainingdeepopticalflow}
    \REQUIRE $\newline \text{Img}_t \text{: Observed image from onboard camera at timestep $t$, } \newline \text{W, H: Image width, height}, \mathbf{X}_{\text{robot},t} \text{: Robot states at timestep $t$}, \newline \mathbf{q} \text{: Robot orientation,} ~\text{MPC} \text{: Model predictive optimal controller, } \newline J_{\text{robot}}(\mathbf{X}_{\text{robot}}) \text{: Task-dependent state cost function for MPC,} \newline $$f_{\text{robot}}(\mathbf{X}_{\text{robot}}, \mathbf{U}) \text{: Robot Dynamics, } \newline \text{OptFlow(Img}$$_t, \text{Img}_{t+1} \text{): Function calculating optical flow,} \newline \mathbf{N}_{data} \text{: Number of data points for training, } \newline \textcolor{black}{ \mathbf{N}_{epoch} \text{: Number of training epochs, }} \newline \textcolor{black}{\mathbf{N}_{batch} \text{: Number of batches in total data, } } \newline \phi \text{: Initial weights and biases of } DOF \text{ NN, } \newline Adam \text{: Stochastic optimization algorithm \cite{adam}}$
\FOR{$t = 1 : \mathbf{N}_{data}$}
\STATE $U^*_t \leftarrow $MPC$(J_{\text{robot}}\text{(}\mathbf{X}_{\text{robot},t}\text{)}, f_{\text{robot}}(\mathbf{X}_{\text{robot},t}, \mathbf{U}_t))$
\STATE $l_{t-1}, \theta_{t-1} \leftarrow$ OptFlow(Img$_{t-1}$, Img$_t$)
\ENDFOR
\FOR{1 : $\mathbf{N}_{epoch}$}
\FOR{1 : $\mathbf{N}_{batch}$}
\STATE $\mathcal{L} = 0$
\FOR{1 : $\#$ of images in a batch}
\FOR{$u$ = 1 : W}
\FOR{$v$ = 1 : H}
\STATE $\hat{l}, \hat{\theta} \leftarrow DOF(\mathbf{q}, u, v, \mathbf{U})$ \textcolor{gray}{$\%$ per pixel}
\STATE $\mathcal{L}~ += MSE(l(u, v), \hat{l}) $+$ MSE(\theta(u, v), \hat{\theta})$
\ENDFOR
\ENDFOR
\ENDFOR
\STATE $\phi \leftarrow Adam.step(\mathcal{L}, \phi)$
\ENDFOR
\ENDFOR
\end{algorithmic}
\end{algorithm}

\begin{algorithm} [ht]
\caption{Testing Deep Optical Flow (DOF) Dynamics}
\begin{algorithmic}[1] \label{alg:testingdeepopticalflow}
    \REQUIRE $\newline \text{Detector: detects targets on image, } \mathbf{U} \text{: Control candidate, } \newline DOF \text{: Trained DOF dynamics, } \mathbf{q} \text{: Robot orientation,} \newline \text{Img: Observed image from an onboard camera } $
\STATE $u, v \leftarrow$ Detector(Img) \textcolor{gray}{$\%$ center of the object}
\STATE $l, \theta$ $\leftarrow DOF(\mathbf{q}, u, v, \mathbf{U})$
\STATE $\dot{u}$ = $l$ cos($\theta$), $\dot{v}$ = $l$ sin($\theta$)
\end{algorithmic}
\end{algorithm}

\begin{table}[ht]
  \centering
    \begin{tabular}{c|c|c|c|c|c}
    \toprule
    \multicolumn{6}{c}{Runtime [ms]} \\
    \midrule
    & $N_{\text{pixel}}$ & $N_{\text{sample}}$ & $N_{\text{timestep}}$ & $\mu\pm2\sigma$ & max \\
    \midrule
    YOLOv3 & 204$\times$153 & 1 & 1 & 14.9 $\pm$ 5.6 & 21.1 \\
    \midrule
    \midrule
    DOF & 1$\times$1 & 1 & 1 & 1.5 $\pm$ 0.4 & 2.1 \\
    \midrule
    DOF & 1$\times$1 & 1 & 80 & 6.7 $\pm$ 2.9 & 10.1 \\
    \midrule
    DOF & 1$\times$1 & 512 & 1 & 1.6 $\pm$ 0.7 & 2.8 \\
    \midrule
    DOF & \textbf{1$\times$1} & \textbf{512} & \textbf{80} & \textbf{8.6 $\pm$ 3.0} & \textbf{11.9} \\
    \midrule
    DOF & 204$\times$153 & 1 & 1 & 6.9 $\pm$ 1.5 & 9.4 \\
    \midrule
    DOF & 204$\times$153 & 1 & 80 & 327.0 $\pm$ 10.6 & 340.3 \\
    \midrule
    DOF & 204$\times$153 & 512 & 1 & OOM & OOM \\
    \midrule
    SpyNet & 192$\times$160 & 1 & 1 & 3.3 $\pm$ 0.5 & 4.0 \\
    \midrule
    SpyNet & 192$\times$160 & 512 & 1 & OOM & OOM \\
    \bottomrule
    \end{tabular}
  \caption{\textcolor{black}{The runtime comparison of our DOF dynamics NN and the state-of-the-art whole-image based optical flow prediction NN, the SpyNet \cite{Ranjan2017SpyNet}. $N_{\text{sample}}$ is the number of samples (batch) used in a sampling-based controller, MPPI, to propagate in parallel with GPU and $N_{\text{timestep}}$ is for multi-step prediction (MPC), which requires sequential computation. OOM (Out of Memory) shows that the full-size optical flow prediction cannot be run in some cases. The runtime was measured with Intel Xeon(R) CPU E5-1650 v4 @ 3.60GHz x 12 CPU and NVIDIA GeForce GTX 1060 6GB.}}
  \label{table:runtime}
\end{table}
We have included a comparison table, \cref{table:runtime}, that shows the differences in runtimes of our optical flow prediction with another state-of-the-art network. However, even though our DOF dynamics approach cares about accuracy, our primary constraint was speed. Therefore, we compared our network with the fastest and smallest of state-of-the-art networks, the SpyNet\cite{Ranjan2017SpyNet}. 
We refer to Table 9 in \cite{Ilg2018ECCV} for benchmark results for optical flow.
The table shows the accuracy and the runtime of the state-of-the-art approaches (\cite{flownet2, Sun_2018_CVPR, Ranjan2017SpyNet}, etc).
%It has the least number of parameters among the state-of-the-art optical flow NN architectures and has the fastest reported runtime.
Note that the total number of parameters in DOF dynamics NN is 34,690, whereas the SpyNet has 1,200,250 parameters.
We tested DOF dynamics both with a single pixel prediction case and the whole-image prediction case (31,212 pixels).
We clearly see that for multi-step prediction ($N_{\text{timestep}}=80$), running DOF dynamics for a whole image ($N_{\text{pixel}}=204\times153$) to predict the optical flow is too slow (3 Hz) and does not fit into real-time MPC algorithms.
Since the SpyNet requires the input image pairs to have width and height to be multiple of 32, we resized the image to have a similar size as our training data set: 192$\times$160=30,720 pixels.
From \cref{table:runtime}, it is apparent that the single pixel approach with our DOF dynamics can only fit into a real-time ``sampling-based'' MPC framework.

We believe comparing the accuracy of our method and the standard full-image optical flow method is unfair because both approaches use different information to predict the optical flow. While the full-image approach uses more perceptual information, our DOF approach uses more non-visual information; the robot orientation and controls.

\textcolor{black}{We report the prediction error of our DOF dynamics on the test dataset in Average Endpoint Error (AEE) of \textbf{2.45}. The endpoint error calculates the Euclidean distance between the ground truth optical flow vectors and the predicted vectors. In the optical flow literature, depending on the training dataset, the state-of-the-art methods report AEE of 0.5-10.0.}

\section{Pixel Model Predictive Control}
\label{sec:PixelMPC}
In this chapter, we introduce Pixel Model Predictive Control (PixelMPC) algorithm for visual object tracking and autonomous racing. PixelMPC literally predicts the future state trajectory of a ``pixel model'', the deep optical flow (DOF) dynamics, and calculates the optimal control sequence (\cref{fig:pixeltraj}).

Assuming we have a visual object detector, for example, detecting custom classes of objects using You Look Only Once (YOLO) \cite{yolov3} algorithm. Given some detected objects, we can predict the future trajectories of their center points/pixels [$u, v$]. For a visual tracking task, one cost function for the optimal control of the DOF dynamics can be the L1 distance between the object pixel position [$u, v$] and the center of the image. This L1 cost function will force the pixel to be close to the center of the image $O$:
\begin{align}\label{eq:pixelcost}
    J_{\text{pixel}}(\mathbf{X}_{\text{pixel}}) = \int_{t=0}^{t_f} c_{\text{pixel}}L1([u_t, v_t], O) dt.
\end{align}
This cost function is reasonable for visual object tracking because the closer the target is to the center of the image, the longer we observe the target. In addition, the center of the image has the least distortion, which means the lowest information lost.

\begin{figure}[ht]
\includegraphics[width=0.49\textwidth]{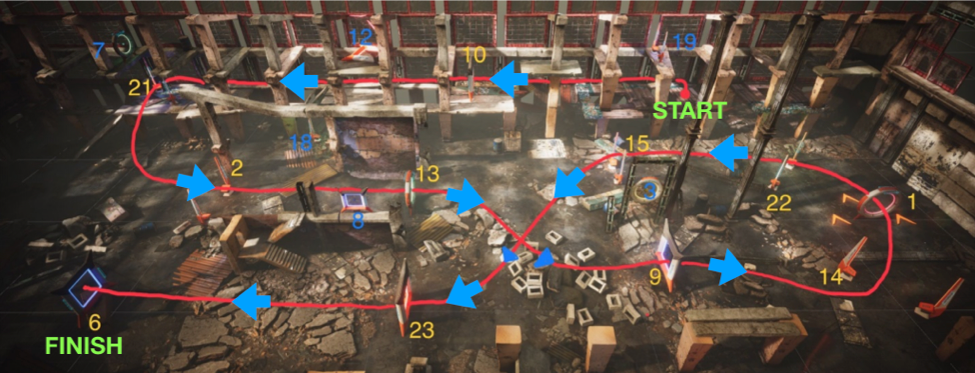}\caption{The race course in the FlightGoggles \cite{FlightGoggles} used in this paper. Image credit: the AlphaPilot--Lockheed Martin AI Drone Racing Innovation Challenge\textsuperscript{\protect\ref{footnote:alphapilot}}.}\label{fig:course}
\end{figure}

The autonomous racing task-related cost function for a finite-horizon optimal control problem can be designed in a form of \cref{eq:costfunction}. For example, to follow the desired position, orientation, and velocity $\mathbf{p}_d, \mathbf{q}_d,$ and $\mathbf{v}_d$:
\begin{align}
    J_{\text{robot}}(\mathbf{X}_{\text{robot}}) &= \int_{t=0}^{t_f} c_1h(\mathbf{p}_d, \mathbf{p}_t)^2 + c_2(\mathbf{q}_d-\mathbf{q}_t)^2 \nonumber \\
    &\hspace{1.2cm}+ c_3(\mathbf{v}_d-\mathbf{v}_t)^2 dt,
    \label{eq:statecost}
\end{align}
which control cost is ignored and $h(\mathbf{p}_d, \mathbf{p})$ is an indicator function which returns 1,000 if a robot crashes into a gate or a value between [-1, 1]. A smaller return represents the robot being closer to the desired path. The ordered waypoints (gates) are assumed to be given with the map of the entire racing track (e.g. \cref{fig:course}). \textcolor{black}{Note that this information including the position of the targets is only for the racing task along with a real-time path planning, not for the visual-servoing task. If the task of PixelMPC is similar to \cite{falanga2018pampc}, where the task is following given waypoints, then prior information of target locations is not required.}

Now, the total cost function for the optimization \cref{eq:optcontrolprob} is formed as
\begin{align}\label{eq:totalJ}
    J(\mathbf{X}) = J_{\text{robot}}(\mathbf{X}_{\text{robot}}) + J_{\text{pixel}}(\mathbf{X}_{\text{pixel}}),
\end{align}
where a new state $\mathbf{X}$ is defined as $\mathbf{X}=[\mathbf{X}_{\text{robot}}; \mathbf{X}_{\text{pixel}}]=[\mathbf{p}, \mathbf{q}, \mathbf{v}, u; v] = [x, y, z, q_w, q_x, q_y, q_z, \dot{x}, \dot{y}, \dot{z}, u, v]^T$.

The total dynamics $F(\mathbf{X}, \mathbf{U})$ used to optimize \cref{eq:totalJ} can be written as a combination of two dynamics Eqs. (\ref{eq:dronedynamics1})-(\ref{eq:dronedynamics4}) and \cref{eq:DOFdynamics}. % The total dynamics are also described in \cref{fig:fx}.
% Careful cost function design is needed to handle the conflicting tasks of high-speed flying and visual tracking.
Our formulation allows us to emphasize one task over another by tuning the cost function. If we want to achieve a faster speed instead of more visibility, then we can weight it more heavily.
% The generality of the cost function is courtesy of MPPI.

\textcolor{black}{
\cref{alg:PixelMPC} shows the PixelMPC algorithm. Either from a ground truth or a state estimator, we receive a new robot state and an image from a monocular camera. A detector (e.g. YOLOv3 \cite{yolov3}) detects the center of a target (gate, in a racing scenario) $(u, v)$ on the image space and an optimal model predictive controller solves the optimization problem with respect to the total cost $J$, \cref{eq:totalJ}, with a receding time horizon $\mathbf{T}$. After propagating the combined model dynamics and running the optimization step, we execute the first control action and use the remaining control trajectory solution for the next optimization loop as a warm start. Then again we receive a new robot state with an image and repeat the optimization at a rate of 40 Hz.
}

\begin{algorithm} [ht]
\caption{Pixel Model Predictive Control (PixelMPC)}
\begin{algorithmic}[1] \label{alg:PixelMPC}
    \REQUIRE $\newline \text{Detector \cite{yolov3}: detects targets on image, } \newline DOF \text{: Deep optical flow dynamics, } \Delta t \text{: timestep size, } \newline \text{Img: Observed image from an onboard camera, } \newline \text{Ctrl$^*(J$($\mathbf{X}, \mathbf{U}$)): Optimal controller, computes } d\mathbf{U}^*, \newline J_{\text{robot}}(\mathbf{X}_{\text{robot}}$$) \text{: Task-dependent robot state cost function, } \newline J_{\text{pixel}}(u, v) \text{: Task-dependent pixel state cost function, } \newline f_{\text{robot}} \text{: Robot dynamics}, \mathbf{U}_{0:T} \text{: Initial control sequence, } \newline \mathbf{T} \text{: MPC time horizon, } \mathbf{K} \text{: Number of optimization}$
\WHILE{Task done}
\STATE Receive a new state $\mathbf{X}_{\text{robot},0}$ and Img
\FOR{$k = 0 : \mathbf{K}$}
\STATE $u_0, v_0 \leftarrow$ Detector(Img) \textcolor{gray}{$\%$ center of the object}
\FOR{$t = 0 : \mathbf{T}$}
\STATE $J_t$ = $J_{\text{robot}}(\mathbf{X}_{\text{robot},t})$ + $J_{\text{pixel}}(u_t, v_t)$
\STATE $\mathbf{X}_{\text{robot},t+1} = \mathbf{X}_{\text{robot},t} + f_{\text{robot}}(\mathbf{X}_{\text{robot}, t}, \mathbf{U}_t)\Delta t$ \\
$l, \theta \leftarrow DOF(\mathbf{q}_t, u_t, v_t, \mathbf{U}_t)$ \\
$\dot{u} = l$ cos($\theta$), $\dot{v} = l$ sin($\theta$) \\
$u_{t+1} = u_t + \dot{u}\Delta t, v_{t+1} = v_t + \dot{v}\Delta t$
\ENDFOR
\STATE $d\mathbf{U}^*_{\textcolor{black}{0}:T} \leftarrow$ Ctrl$^*$\Big($ J_{0:T}(\mathbf{X}_{\text{robot},0:T}, u_{0:T}, v_{0:T}, \mathbf{U}_{0:T})$\Big)
\STATE $\mathbf{U}_{\textcolor{black}{0}:T} \leftarrow \mathbf{U}_{\textcolor{black}{0}:T} + d\mathbf{U}^*_{\textcolor{black}{0}:T}$
\ENDFOR
\STATE Execute $\mathbf{U}_0$
\STATE $\mathbf{U}_{0:T-1} \leftarrow \mathbf{U}_{1:T}$ \textcolor{gray}{$\%$ shift for a warm-start}
\ENDWHILE
\end{algorithmic}
\end{algorithm}

% \begin{figure*}
% \centering
% \begin{minipage}[b]{.28\textwidth}
% \includegraphics[width=1.0\textwidth]{figs/course.png}\caption{The race course in the FlightGoggles \cite{FlightGoggles}.}\label{fig:course}
% \end{minipage}
% \quad
% \begin{minipage}[b]{.69\textwidth}
% \includegraphics[width=1.0\textwidth]{figs/block.pdf}\caption{A block diagram showing the algorithmic flow of PixelMPC.}\label{fig:block}
% \end{minipage}
% \end{figure*}

\begin{figure*}[ht]
    \centering
    \includegraphics[width=0.99\textwidth]{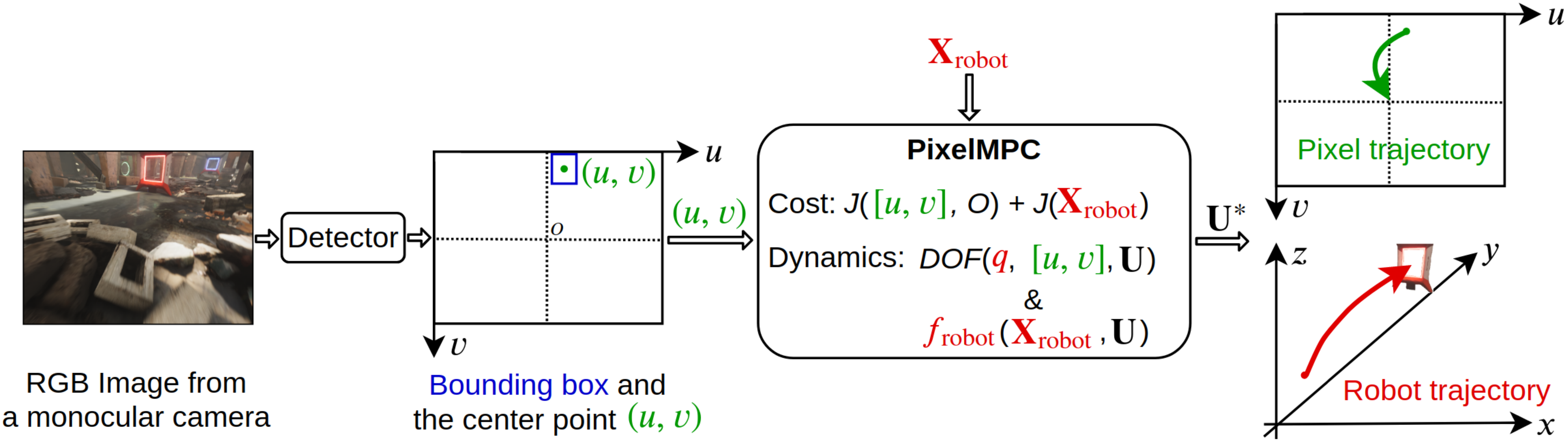}
    \caption{A block diagram showing the algorithmic flow of PixelMPC.}\label{fig:block}
\end{figure*}

\section{Experiments/Results}
\label{sec:Experiments}
\subsection{Experimental Setup}
We tested our algorithm in the FlightGoggles simulation \cite{FlightGoggles}, which is developed for agile flight simulation with high fidelity. The racing scenario is from the AlphaPilot--Lockheed Martin AI Drone Racing Innovation Challenge\footnote{https://www.herox.com/alphapilot/85-2019-virtual-qualifier-tests \label{footnote:alphapilot}} (\cref{fig:course}).

% \begin{figure}[ht]
%     \centering
%     \includegraphics[width=0.3\textwidth]{figs/course.png}
%     \caption{The race course in the FlightGoggles simulation \cite{FlightGoggles} used in this paper. Image credits to the AlphaPilot--Lockheed Martin AI Drone Racing Innovation Challenge webpage\textsuperscript{\protect\ref{footnote:alphapilot}}. } \label{fig:course}
%     \vspace{-0.25cm}
% \end{figure}

We used the quadrotor's dynamics model introduced in \cref{sec:dronedynamics} Eqs. (\ref{eq:dronedynamics1})-(\ref{eq:dronedynamics4}).

To derive our DOF dynamics from optical flow data, we collected 10 rounds of autonomous flight using a nominal MPPI controller, which took around 30 seconds for each round. To fully explore the state space we varied the target speed between 6m/s and 14m/s across rounds. The timestep in MPPI was 0.025 seconds. In total 14,000 images from a monocular camera along with drone states and controls were collected. The images were each downsized to a size of [204, 153]. This provided 204$\times$153=31,212 data points. As a result, around 437 million data points for training DOF were collected from 5 minutes of flying data. The states are collected from ground truth provided in the FlightGoggles simulator.

For object detection, we used one of the state-of-the-art algorithms, YOLOv3 \cite{yolov3}, which allows us to detect multiple objects in real-time. 3,000 downsized images were used to train the YOLOv3 model to predict 7 classes of gates in the FlightGoggles racing environment (\cref{fig:course}).
% All the experiments were done using Intel Xeon(R) CPU E5-1650 v4 @ 3.60GHz x 12 CPU and NVIDIA GeForce GTX 1060 6GB.
% The PixelMPC results can be also found in the supplementary video.

\subsection{Model Predictive Path Integral control (MPPI)}
In this work, out of many real-time MPC algorithms, we adopt the sampling-based stochastic optimal control algorithm, the Model Predictive Path Integral control (MPPI) \cite{mppi}. MPPI allows us to handle stochasticity and it provides the easiness of designing and tuning non-quadratic cost functions, compared to other optimal control algorithms where most of them require a quadratic cost function.

For a drone racing task along with the visual object tracking task, the cost function parameters used for MPPI are $\Delta t$=0.025 ($sec$), $c_1$=400, $c_2$=250, $c_3$=8.0, $T$=80, and $K$=1. The control variance had noise profiles: $\sigma_{\text{roll}}$=0.2, $\sigma_{\text{pitch}}$=0.2, $\sigma_{\text{yaw}}$=0.3, and $\sigma_{\text{thrust}}$=2.2.
\textcolor{black}{$c_{\text{pixel}}$ was tuned between 3.0e+6 and 9.0e+6 and the resulting different behaviors are reported in \cref{table:lostcount} and \cref{table:laptime}. The reason why the parameter $c_{\text{pixel}}$ is chosen 4 orders of magnitude higher than all other cost function parameters is because we only normalized the pixel position term from [0, W]$\times$[0, H] to [0.0, 1.0]$\times$[0.0, 1.0].}
A total of 512 samples were used to propagate the 12 states with a time horizon of 80 in 40 Hz, which results in a 2 second trajectory. The number of samples depends on the hardware (GPU, CPU, RAM, etc.) and the size of the DOF NN dynamics.
\textcolor{black}{The nominal MPPI case for the racing task used the cost function only composed of \cref{eq:statecost} and the same parameters described above were used to give a fair comparison.}

\textcolor{black}{Although the drone dynamics we introduced in this paper are the simplified linear dynamics from the simulator \cite{FlightGoggles}, any nonlinear dynamics model can also be used as robot dynamics model in PixelMPC.} %This is due to the formulation of underlying controller MPPI.

\subsection{Drone Racing with Object Tracking}
We compare the visibility in percentage; how long the robot grabs the target in its view.
In the PixelMPC framework, there are some additional DOF dynamics-related parameters we can tune: 1) the time horizon $t_f$ considered for the pixel cost and 2) the cost coefficient $c_{\text{pixel}}$. $t_f=1.0 sec$ means the pixel cost \cref{eq:pixelcost} only penalizes the pixel trajectory within 1.0 second.

\cref{table:lostcount} shows the time the drone loses the target in $sec$ ($\mu\pm2\sigma$). We consider the `loss' as visually losing the target after the robot first sees it. In this experiment, we used the ground truth provided by the FlightGoggles for robot states.

\begin{figure}[t]
  \centering
  \includegraphics[width=0.49\textwidth]{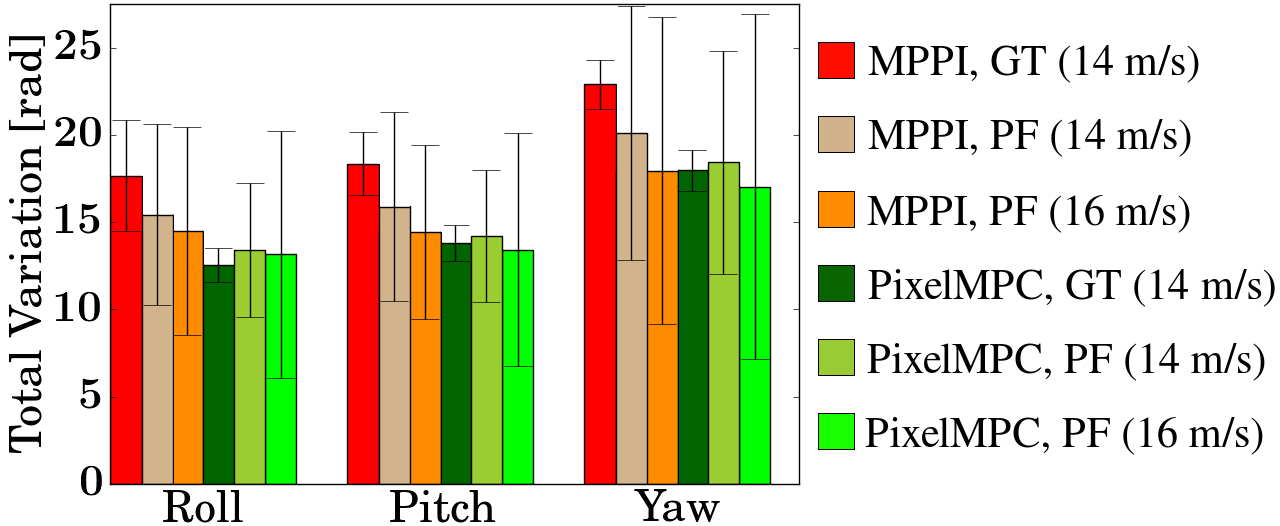}
  \caption{The total variation of roll, pitch, yaw angles of 25 laps of robot trajectories running the nominal MPPI and the PixelMPC. Both controllers were tested with ground truth (GT) and a particle filter (PF) state estimator. The error bar represents $\mu\pm2\sigma$. The smaller total variation of the robot orientation implies less shaky robot behavior.}
  \label{fig:total_variation}
\end{figure}

In the nominal case, without considering DOF dynamics in MPPI, the time the robot has less than 50\% visibility of a target was .6 $sec$, which is more than 42\% of the total flying time ($\sim$31.8 $sec$). With PixelMPC, we can decrease it to 1.5 $sec$, 4.5\% of the flying time ($\sim$33.5 $sec$).
The time of robot having less than 0\% visibility of a target also decreased from 3.6 $sec$ (4\% of flying time) to 0.2 $sec$ (less than 1\% of flying time).
Notice that in both 0\% and 50\% cases, the 2$\sigma$ of the lost time is very large in the nominal case, compared to the PixelMPC cases. This can be explained in \cref{fig:total_variation}, where the plots show how smooth the movement is when we use PixelMPC.
\textcolor{black}{Also, compared to the nominal MPPI, PixelMPC showed 29$\%$ decrease in linear and angular accelerations in mean, which resulted in a slower speed but it provided much smoother behavior; please see \cref{fig:total_variation} (Best shown in the supplementary video). However, the smoothness behavior of PixelMPC is a byproduct of the visual target tracking, not the main goal. Also, the visual target tracking cannot be accomplished by simply applying a smoothing/filtering to a controller.}

\begin{table}
  \centering
    \begin{tabular}{c|*{4}r}
    \toprule
    \multicolumn{5}{c}{Time ($sec$) of less than 50\% visibility} \\
    \midrule
    \diagbox{$t_f$}{$c_{\text{pixel}}$} & 0.0 & 3.0e+6 & 6.0e+6 & 9.0e+6 \\
    \midrule
    0.0 & 13.6$\pm$3.6 & - & - & - \\
    % 0.025 & - & 7.3$\pm$0.6 & 7.5$\pm$1.0 & 8.5$\pm$1.3 \\
    1.0 & - & 3.6$\pm$0.6 & 1.9$\pm$0.6 & \textbf{1.5$\pm$0.2} \\
    2.0 & - & 3.1$\pm$0.9 & 2.0$\pm$0.6 & 1.9$\pm$1.2 \\
    \end{tabular}
    \begin{tabular}{c|*{4}r}
    \toprule
    \multicolumn{5}{c}{Time ($sec$) of 0\% visibility} \\
    \midrule
    \diagbox{$t_f$}{$c_{\text{pixel}}$} & 0.0 & 3.0e+6 & 6.0e+6 & 9.0e+6 \\
    \midrule
    0.0 & 3.6$\pm$1.1 & - & - & - \\
    % 0.025 & - &2.3$\pm$0.4 & 3.6$\pm$0.4 & 2.5$\pm$0.3 \\
    1.0 & - & 1.0$\pm$0.3 & 1.1$\pm$0.5 & \textbf{0.2$\pm$0.1} \\
    2.0 & - & 0.7$\pm$0.4 & 0.6$\pm$0.2 & \textbf{0.2$\pm$0.1} \\
    \bottomrule
    \end{tabular}
    \caption{The time ($sec$) the robot visually loses the target. The results are from 10 races per each case in $\mu\pm2\sigma$ v.s. [$t_f (sec), c_{\text{pixel}}$] when the target speed was 14$m/s$. The top-left [$t_f, c_{\text{pixel}}$]=[0.0, 0.0] shows the nominal MPPI running without any DOF dynamics control.}
  \label{table:lostcount}
\end{table}

In \cref{table:laptime}, we compare the race time for each case to see how much lap time delay we get to pay for more visual information. \cref{table:laptime} shows the mean and the 2$\sigma$ standard deviation from 10 laps of racing per each case. As expected, the PixelMPC loses lap time by achieving more visibility of the racecourse. However, we believe it is worth to pay 1.7 $sec$, sometimes less than 0.2 $sec$, to achieve 42\% $\rightarrow$ 4.5\% decrease in time that the robot loses important information in its view.

\begin{figure}
\begin{subfigure}{.22\textwidth}
  \centering
  \vspace{-0.2cm}
  \includegraphics[width=\textwidth]{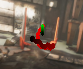}
\end{subfigure}
% \vspace{2mm}
\begin{subfigure}{.26\textwidth}
  \centering
  \vspace{-0.4cm}
  \includegraphics[width=\textwidth]{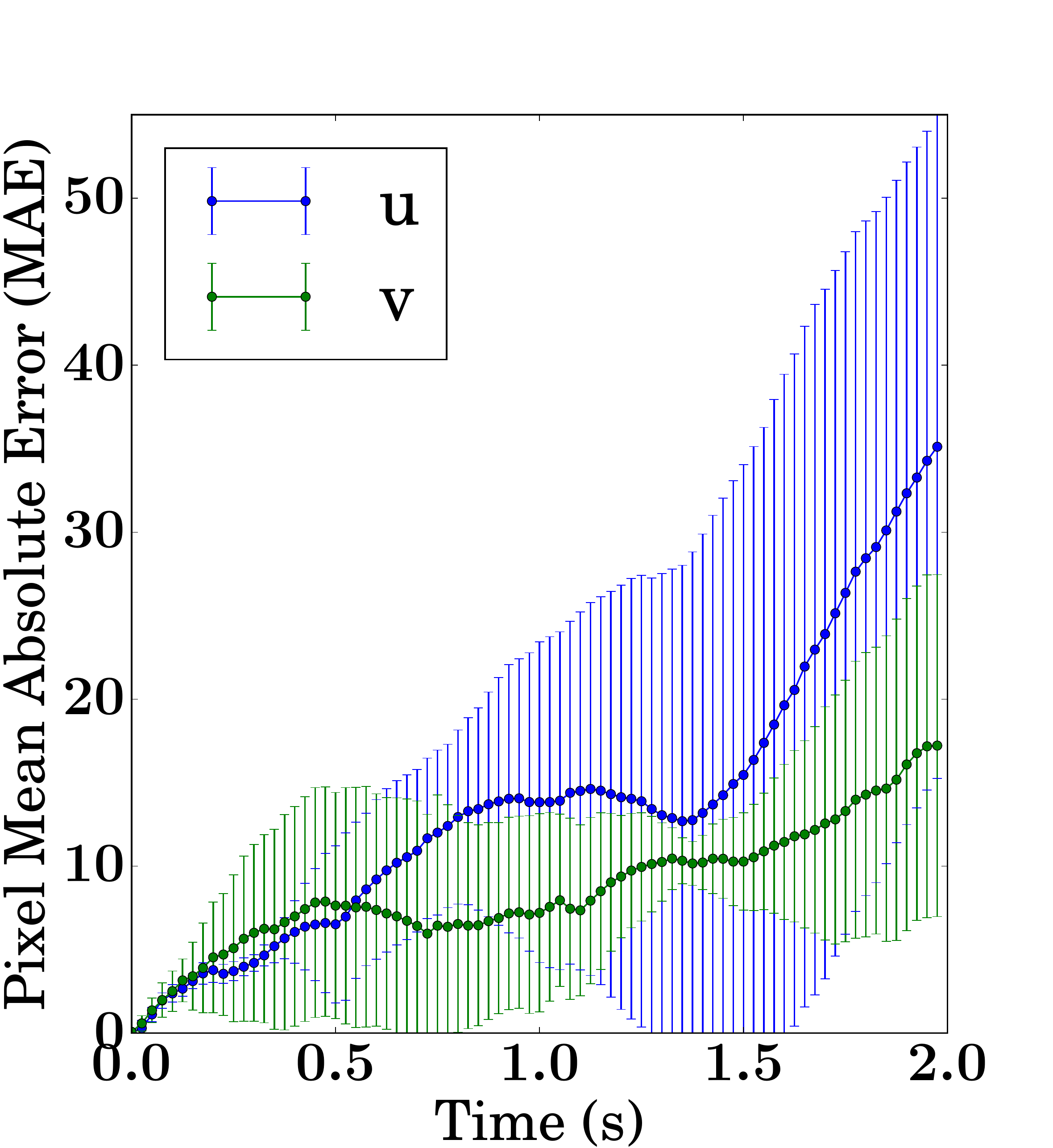}
\end{subfigure}
\caption{\textcolor{black}{\textit{Left}: Cropped image showing PixelMPC-predicted pixel trajectory (Green) vs. Actual pixel trajectory (Red). \textit{Right}: Mean absolute error and standard deviation of the multi-step prediction of pixel position on DOF dynamics.} }
\label{fig:multistep_mae}
% \vspace{-0.5cm}
\end{figure}

\textcolor{black}{We also report DOF dynamics' multi-step prediction error in \cref{fig:multistep_mae}. We see the predicted pixel trajectory is shorter than the actual trajectory in general, but the predicted trajectory closely follows the actual trajectory direction-wise. Note that our MPC scheme solves this compounding error problem with feedback and real-time optimization. }

\begin{table}
  \centering
    \begin{tabular}{c|*{4}r}
    \toprule
    \multicolumn{5}{c}{Lap time ($sec$)} \\
    \midrule
    \diagbox{$t_f$}{$c_{\text{pixel}}$} & 0.0 & 3.0e+6 & 6.0e+6 & 9.0e+6 \\
    \midrule
    0.0 & \textbf{31.8$\pm$1.0} & - & - & - \\
    % 0.025 & - & 32.0$\pm$0.7 & 32.1$\pm$0.4 & \textbf{31.7$\pm$0.4} \\
    1.0 & - & 32.7$\pm$0.4 & 33.2$\pm$0.2 & 33.5$\pm$0.2 \\
    2.0 & - & 33.2$\pm$0.1 & 34.2$\pm$0.3 & 34.0$\pm$0.5 \\
    \bottomrule
    \end{tabular}
  \caption{Lap time from 10 races per each case in $\mu\pm2\sigma$ ($sec$) v.s. [$t_f (sec), c_{\text{pixel}}$] when the target speed is 14$m/s$. % The top-left lap time [$t_f, c_{\text{pixel}}$]=[0.0, 0.0] shows the nominal MPPI running without any DOF dynamics control.
  }
  \label{table:laptime}
  \vspace{-0.2cm}
\end{table}

\begin{figure*}[ht]
  \centering
  \includegraphics[width=0.99\textwidth]{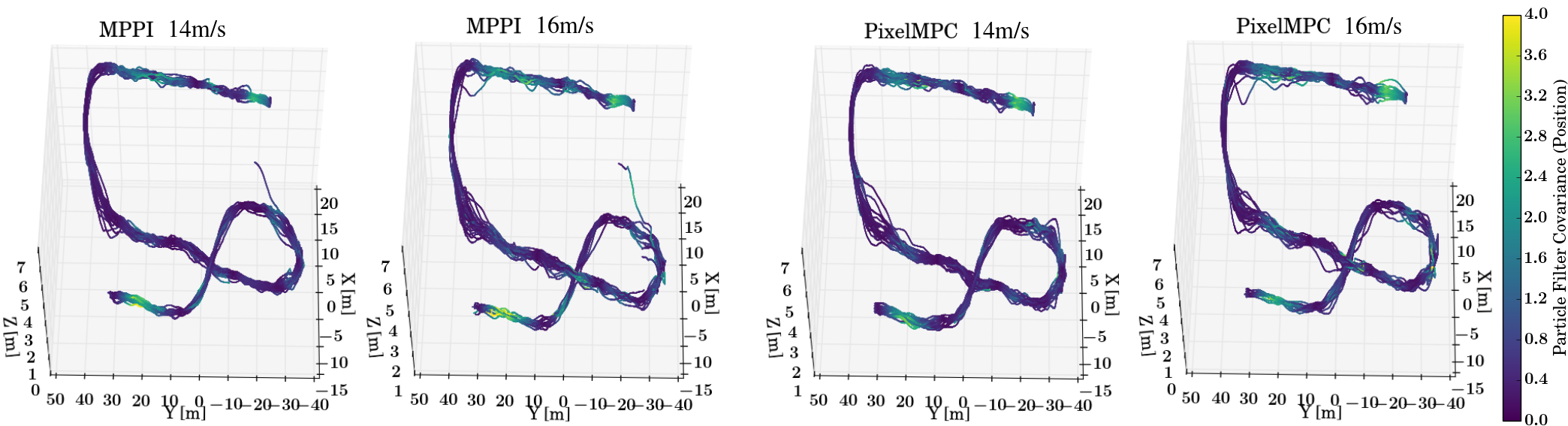}
  \caption{25 Laps of running the nominal MPPI (\textit{Left}) and the PixelMPC ($c_{\text{pixel}}$=9.0e+6 and $t_f$=1.0) (\textit{Right}) with a particle filter. The target speed was set to 14 $m/s$ and 16 $m/s$. The color represents the total covariance in position. }
  \label{fig:robot_trajs}
\end{figure*}

\subsection{Vision-based State Estimation with Particle Filter}
\label{sec:state_estimation}
For state estimation with sensors (IMU, cameras, etc.), having more visual information and smooth flying behavior will benefit the state estimation and result in fewer failures overall. The most likely cause of a collision is an inaccurate state estimate.
%In other words, the robot will collide less, as the collision mostly comes from unreliable state estimation.
% Now, to be more realistic, let us assume that we do not have a ground truth of a robot's state.
In a racing scenario, we can still assume that the racing map, i.e. the gates' location information is given. Then, one of the biggest challenges will be estimating the robot's state, to perform accurate path planning and control.

For estimating the robot's state, we use a particle filter with an observation model using gate information from observed images. The particle filter is run with 6400 particles and uses the GPU to parallelize the motion and sensor updates.

\subsubsection{Motion Update}

The motion update of the particle filter is done by integrating the IMU measurements directly. Additional Gaussian noise is injected into the filter with mean $0$ and variance $0.2$ directly on position [$m$]. In addition to that, Gaussian noise is added to the IMU measurements directly both with mean $0$ and variance $0.2$ for acceleration and variance $0.1$ for angular rates. These tunings allow the particle filter to quickly jump to whatever sensor update occurs, but make the state estimate very unstable. The filter's covariance will quickly balloon without any feature detections.

\subsubsection{Sensor Update}

The only sensor model of the particle filter is to use the nominal locations of the gate corners in the 3D world and back project them into the image plane. Then we find the difference between the detected results and the expected ones. Any missing detection is penalized heavily by $4\times$ W where W is the width of the camera image. Our custom YOLOv3 \cite{yolov3} gate detector is used to generate the detection of the 2D positions from an image along with a bounding box, which includes the third (depth) information.

\cref{table:successrate} shows that, with the target speed of 14 $m/s$, the success rate of both cases are the same (80$\%$) but if we increase the target speed to 16 $m/s$ with the same cost parameters, the PixelMPC reports a higher success rate. \textcolor{black}{The failure (crash) cases came from losing target visibility which resulted in the divergence of the state estimation.} The 25 trajectories of running PixelMPC ($t_f$=1.0, $c_{pixel}$=9.0e+6) and the nominal MPPI with a particle filter is shown in \cref{fig:robot_trajs}.
Since the racetrack we used only allows few seconds of flying between two consecutive gates, it is not intuitive to see if the PixelMPC decreases the particle filter covariance because even nominal MPPI could see the target gates very often. Therefore, we did one more straight-line flying test to fully see the effect of PixelMPC on state estimation. In this case, we increased the target speed to 20 $m/s$, where MPPI has to pitch down a lot to hit the target speed. As soon as the detector detects the gates, the PixelMPC tries to grab the feature in its view and this results in a smaller covariance of the particle filter. 
\textcolor{black}{The last column of \cref{table:successrate} shows the maximum covariance of position from 25 runs of nominal MPPI and PixelMPC.}

\begin{table}[ht]
  \centering
    \begin{tabular}{c|*{2}c|c}
    \toprule
    & \multicolumn{2}{c}{[Success rate ($\%$), Lap time ($sec$)]} & $\Sigma_{max}$\\
    \midrule
    \diagbox{Ctrl}{$\mathbf{v}_{d}$} & 14$m/s$ & 16$m/s$ & 20$m/s$\\
    \midrule
    MPPI & [80$\%$, 31.8$\pm$0.7] & [52$\%$, 29.6$\pm$0.8] & 9.2 \\
    PixelMPC & [80$\%$, 33.0$\pm$0.8] & [\textbf{60}$\%$, 30.6$\pm$0.7] & \textbf{5.7} \\
    \bottomrule
    \end{tabular}
  \caption{Success rate ($\%$) of 25 laps (14 $m/s$ and 16 $m/s$) and the success case lap time ($sec$) of running MPPI and PixelMPC ($t_f$=1.0, $c_{\text{pixel}}$=9.0e+6) with a particle filter. The maximum covariance of position $\Sigma_{max}$ was tested with a target speed 20 $m/s$ on a straight lane.}
  \label{table:successrate}
\end{table}

% \begin{table}[ht]
%   \centering
%     \begin{tabular}{p{2cm}|p{2cm}}
%     \toprule
%     \multicolumn{2}{c}{Maximum covariance in position [m]} \\
%     \midrule
%     MPPI & PixelMPC \\
%     \midrule
%     9.2 & \textbf{5.7} \\
%     \bottomrule
%     \end{tabular}
%   \caption{The maximum covariance of position from 25 laps of running MPPI and PixelMPC ($t_f$=1.0, $c_{\text{pixel}}$=9.0e+6) on a straight lane with a particle filter. The target speed was 20 $m/s$.}
%   \label{table:straight_cov}
% \end{table}

% \begin{figure}[ht]
%   \centering
%   \includegraphics[width=0.35\textwidth]{figs/straight_cov.png}
%   \caption{25 runs of MPPI and PixelMPC with a particle filter at the first straight line. The target speed was set to 20 $m/s$ and we can see the decreased covariance in the PixelMPC case, especially between the first gate and the second gate.}
%   \label{fig:straight_cov}
% \end{figure}

\section{Conclusion}
\label{sec:conclusion}
By fusing vision, path planning, and control into a single optimization framework, high-speed racing can be accomplished with more stable state estimation along with more visual information.
Our algorithm can be generally used in any camera-based robot system for visual servoing.
% \textcolor{black}{Also, if we invert the pixel cost function above, PixelMPC can achieve the obstacle avoidance task as well.}
\textcolor{black}{Testing our algorithm with real hardware will be our next step to move forward, but there is still room for improvement.}
The suggested deep optical flow (DOF) dynamics does not take the depth/distance of the target pixel \textcolor{black}{and the robot's velocity} information into account. The current DOF approach works well thanks to the generalization property of the deep neural network, but incorporating the target pixel's depth information will result in a more robust dynamics propagation. Another direction to robustify the suggested dynamics will be propagating the target bounding box, i.e. the 4 corners of it, like a particle filter approach.
% This approach will also help to robustify the suggested DOF dynamics.
\textcolor{black}{Lastly, although the constant target velocity settings for racing and other inputs indirectly include the velocity information, directly incorporating the velocity will be helpful also for other non-racing tasks dealing with variable speed and other specific maneuvers.}

% \appendices

% \section*{Acknowledgment}
% This research was supported by NASA.

\bibliographystyle{IEEEtran}
\bibliography{IEEEabrv,pixelmpc}

\begin{thebibliography}{10}
\providecommand{\url}[1]{#1}
\csname url@samestyle\endcsname
\providecommand{\newblock}{\relax}
\providecommand{\bibinfo}[2]{#2}
\providecommand{\BIBentrySTDinterwordspacing}{\spaceskip=0pt\relax}
\providecommand{\BIBentryALTinterwordstretchfactor}{4}
\providecommand{\BIBentryALTinterwordspacing}{\spaceskip=\fontdimen2\font plus
\BIBentryALTinterwordstretchfactor\fontdimen3\font minus
  \fontdimen4\font\relax}
\providecommand{\BIBforeignlanguage}[2]{{%
\expandafter\ifx\csname l@#1\endcsname\relax
\typeout{** WARNING: IEEEtran.bst: No hyphenation pattern has been}%
\typeout{** loaded for the language `#1'. Using the pattern for}%
\typeout{** the default language instead.}%
\else
\language=\csname l@#1\endcsname
\fi
#2}}
\providecommand{\BIBdecl}{\relax}
\BIBdecl

\bibitem{PanRSS18}
\BIBentryALTinterwordspacing
Y.~Pan, C.-A. Cheng, K.~Saigol, K.~Lee, X.~Yan, E.~A. Theodorou, and B.~Boots,
  ``\href{http://www.roboticsproceedings.org/rss14/p56.pdf}{Agile Autonomous
  Driving using End-to-End Deep Imitation Learning},'' \emph{Robotics: Science
  and Systems}, 2018. [Online]. Available:
  \url{http://www.roboticsproceedings.org/rss14/p56.pdf}
\BIBentrySTDinterwordspacing

\bibitem{bojarski2016end}
\BIBentryALTinterwordspacing
M.~Bojarski, D.~{Del Testa}, D.~Dworakowski, B.~Firner, B.~Flepp, P.~Goyal,
  L.~D. Jackel, M.~Monfort, U.~Muller, J.~Zhang, X.~Zhang, J.~Zhao, and
  K.~Zieba, ``\href{https://arxiv.org/abs/1604.07316}{End to End Learning for
  Self-Driving Cars},'' apr 2016. [Online]. Available:
  \url{https://arxiv.org/abs/1604.07316}
\BIBentrySTDinterwordspacing

\bibitem{zhang17aaai}
J.~Zhang and K.~Cho, ``Query-efficient imitation learning for end-to-end
  simulated driving,'' in \emph{Thirty-First AAAI Conference on Artificial
  Intelligence}, June 2017.

\bibitem{LevineVisuomotor}
\BIBentryALTinterwordspacing
S.~Levine, C.~Finn, T.~Darrell, and P.~Abbeel,
  ``\href{http://jmlr.org/papers/v17/15-522.html}{End-to-End Training of Deep
  Visuomotor Policies},'' \emph{Journal of Machine Learning Research}, vol.~17,
  no.~39, pp. 1--40, 2016. [Online]. Available:
  \url{http://jmlr.org/papers/v17/15-522.html}
\BIBentrySTDinterwordspacing

\bibitem{giusti2016machine}
A.~Giusti, J.~Guzzi, D.~Ciresan, F.-L. He, J.~P. Rodriguez, F.~Fontana,
  M.~Faessler, C.~Forster, J.~Schmidhuber, G.~Di~Caro, D.~Scaramuzza, and
  L.~Gambardella,
  ``\href{https://ieeexplore.ieee.org/abstract/document/7358076}{A Machine
  Learning Approach to Visual Perception of Forest Trails for Mobile Robots},''
  \emph{IEEE Robotics and Automation Letters}, 2016.

\bibitem{Smolyanskiy2017TowardLA}
N.~Smolyanskiy, A.~Kamenev, J.~Smith, and S.~T. Birchfield,
  ``\href{https://ieeexplore.ieee.org/abstract/document/8206285}{Toward
  low-flying autonomous MAV trail navigation using deep neural networks for
  environmental awareness},'' \emph{2017 IEEE/RSJ International Conference on
  Intelligent Robots and Systems (IROS)}, pp. 4241--4247, 2017.

\bibitem{murORB2}
R.~Mur-Artal and J.~D. Tard\'os,
  ``\href{https://ieeexplore.ieee.org/abstract/document/7946260}{{ORB-SLAM2}:
  an Open-Source {SLAM} System for Monocular, Stereo and {RGB-D} Cameras},''
  \emph{IEEE Transactions on Robotics}, vol.~33, no.~5, pp. 1255--1262, 2017.

\bibitem{Kaufmann19Beauty}
E.~Kaufmann, M.~Gehrig, P.~Foehn, R.~Ranftl, A.~Dosovitskiy, V.~Koltun, and
  D.~Scaramuzza, ``\href{https://ieeexplore.ieee.org/document/8793631}{Beauty
  and the Beast: Optimal Methods Meet Learning for Drone Racing},'' \emph{2019
  IEEE International Conference on Robotics and Automation (ICRA)}, 2019.

\bibitem{Kaufmann18DeepDrone}
E.~Kaufmann, A.~Loquercio, R.~Ranftl, A.~Dosovitskiy, V.~Koltun, and
  D.~Scaramuzza,
  ``\href{http://proceedings.mlr.press/v87/kaufmann18a.html}{Deep Drone Racing:
  Learning Agile Flight in Dynamic Environments},'' in \emph{Proceedings of The
  2nd Conference on Robot Learning}, ser. Proceedings of Machine Learning
  Research, vol.~87.\hskip 1em plus 0.5em minus 0.4em\relax PMLR, 29--31 Oct
  2018, pp. 133--145.

\bibitem{flownet2}
E.~Ilg, N.~Mayer, T.~Saikia, M.~Keuper, A.~Dosovitskiy, and T.~Brox,
  ``\href{http://lmb.informatik.uni-freiburg.de//Publications/2017/IMKDB17}{FlowNet
  2.0: Evolution of Optical Flow Estimation with Deep Networks},'' in
  \emph{IEEE Conference on Computer Vision and Pattern Recognition (CVPR)}, Jul
  2017.

\bibitem{Sun_2018_CVPR}
D.~Sun, X.~Yang, M.-Y. Liu, and J.~Kautz,
  ``\href{http://openaccess.thecvf.com/content_cvpr_2018/html/Sun_PWC-Net_CNNs_for_CVPR_2018_paper.html}{{PWC}-Net:
  CNNs for Optical Flow Using Pyramid, Warping, and Cost Volume},'' in
  \emph{IEEE Conference on Computer Vision and Pattern Recognition (CVPR)},
  June 2018.

\bibitem{Ranjan2017SpyNet}
A.~Ranjan and M.~J. Black.,
  ``\href{http://openaccess.thecvf.com/content_cvpr_2017/html/Ranjan_Optical_Flow_Estimation_CVPR_2017_paper.html}{Optical
  flow estimation using a spatial pyramid network},'' \emph{2017 IEEE
  Conference on Computer Vision and Pattern Recognition (CVPR)}, 2017.

\bibitem{Penin2017MinTime}
B.~{Penin}, R.~{Spica}, P.~R. {Giordano}, and F.~{Chaumette},
  ``\href{https://ieeexplore.ieee.org/abstract/document/8206522}{Vision-based
  minimum-time trajectory generation for a quadrotor {UAV}},'' in \emph{2017
  IEEE/RSJ International Conference on Intelligent Robots and Systems (IROS)},
  Sep. 2017, pp. 6199--6206.

\bibitem{Murali2019DiffFlat}
V.~{Murali}, I.~{Spasojevic}, W.~{Guerra}, and S.~{Karaman},
  ``\href{https://ieeexplore.ieee.org/abstract/document/8814697}{Perception-aware
  trajectory generation for aggressive quadrotor flight using differential
  flatness},'' in \emph{2019 American Control Conference (ACC)}, July 2019, pp.
  3936--3943.

\bibitem{Negeli2017ViewpointOpt}
T.~{Nägeli}, J.~{Alonso-Mora}, A.~{Domahidi}, D.~{Rus}, and O.~{Hilliges},
  ``\href{https://ieeexplore.ieee.org/abstract/document/7847361}{Real-Time
  Motion Planning for Aerial Videography With Dynamic Obstacle Avoidance and
  Viewpoint Optimization},'' \emph{IEEE Robotics and Automation Letters},
  vol.~2, no.~3, pp. 1696--1703, July 2017.

\bibitem{falanga2018pampc}
D.~Falanga, P.~Foehn, P.~Lu, and D.~Scaramuzza,
  ``\href{http://rpg.ifi.uzh.ch/docs/IROS18_Falanga.pdf}{PAMPC:
  Perception-Aware Model Predictive Control for Quadrotors},'' in \emph{2018
  IEEE International Conference on Intelligent Robots and Systems (IROS)},
  2018.

\bibitem{yolov3}
J.~Redmon and A.~Farhadi, ``\href{https://arxiv.org/abs/1804.02767}{YOLOv3: An
  Incremental Improvement},'' \emph{arXiv}, 2018.

\bibitem{fasterrcnn}
S.~Ren, K.~He, R.~Girshick, and J.~Sun,
  ``\href{https://arxiv.org/abs/1506.01497}{Faster R-CNN: Towards Real-Time
  Object Detection with Region Proposal Networks},'' in \emph{Advances in
  Neural Information Processing Systems 28}, C.~Cortes, N.~D. Lawrence, D.~D.
  Lee, M.~Sugiyama, and R.~Garnett, Eds.\hskip 1em plus 0.5em minus 0.4em\relax
  Curran Associates, Inc., 2015, pp. 91--99.

\bibitem{mppi}
G.~{Williams}, N.~{Wagener}, B.~{Goldfain}, P.~{Drews}, J.~M. {Rehg},
  B.~{Boots}, and E.~A. {Theodorou},
  ``\href{https://ieeexplore.ieee.org/abstract/document/7989202}{Information
  theoretic MPC for model-based reinforcement learning},'' in \emph{2017 IEEE
  International Conference on Robotics and Automation (ICRA)}, May 2017, pp.
  1714--1721.

\bibitem{FlightGoggles}
W.~Guerra, E.~Tal, V.~Murali, G.~Ryou, and S.~Karaman,
  ``\href{https://arxiv.org/abs/1905.11377}{FlightGoggles: Photorealistic
  Sensor Simulation for Perception-driven Robotics using Photogrammetry and
  Virtual Reality},'' \emph{arXiv}, 2019.

\bibitem{Farneback03}
G.~Farnebäck,
  ``\href{https://link.springer.com/chapter/10.1007/3-540-45103-X_50}{Two-Frame
  Motion Estimation Based on Polynomial Expansion},'' in \emph{Scandinavian
  Conference on Image Analysis}, vol. 2749, 06 2003, pp. 363--370.

\bibitem{opencv}
G.~Bradski, ``{The OpenCV Library},'' \emph{Dr. Dobb's Journal of Software
  Tools}, 2000.

\bibitem{adam}
D.~P. Kingma and J.~Ba, ``\href{http://arxiv.org/abs/1412.6980}{Adam: {A}
  Method for Stochastic Optimization},'' \emph{Proceedings of the 3rd
  International Conference on Learning Representations (ICLR)}, vol.
  abs/1412.6980, 2014.

\bibitem{tensorflow}
\BIBentryALTinterwordspacing
M.~Abadi, A.~Agarwal, P.~Barham, E.~Brevdo, Z.~Chen, C.~Citro, G.~S. Corrado,
  A.~Davis, J.~Dean, M.~Devin, S.~Ghemawat, I.~Goodfellow, A.~Harp, G.~Irving,
  M.~Isard, Y.~Jia, R.~Jozefowicz, L.~Kaiser, M.~Kudlur, J.~Levenberg,
  D.~Man\'{e}, R.~Monga, S.~Moore, D.~Murray, C.~Olah, M.~Schuster, J.~Shlens,
  B.~Steiner, I.~Sutskever, K.~Talwar, P.~Tucker, V.~Vanhoucke, V.~Vasudevan,
  F.~Vi\'{e}gas, O.~Vinyals, P.~Warden, M.~Wattenberg, M.~Wicke, Y.~Yu, and
  X.~Zheng, ``\href{http://tensorflow.org/}{{TensorFlow}: Large-Scale Machine
  Learning on Heterogeneous Systems},'' 2015, software available from
  tensorflow.org. [Online]. Available: \url{http://tensorflow.org/}
\BIBentrySTDinterwordspacing

\bibitem{Ilg2018ECCV}
E.~Ilg, T.~Saikia, M.~Keuper, and T.~Brox,
  ``\href{http://openaccess.thecvf.com/content_ECCV_2018/papers/Eddy_Ilg_Occlusions_Motion_and_ECCV_2018_paper.pdf}{Occlusions,
  Motion and Depth Boundaries with a Generic Network for Disparity, Optical
  Flow or Scene Flow Estimation},'' \emph{the European Conference on Computer
  Vision (ECCV)}, 2018.

\end{thebibliography}

\onecolumn
\section*{Citations}
\hspace{-0.3cm}Plain Text:
\\ \\
K. Lee, J. Gibson, and E. A. Theodorou, “Aggressive Perception-Aware Navigation using Deep Optical Flow Dynamics and PixelMPC,” in IEEE Robotics and Automation Letters, 2020.
\\ \\
BibTeX:
\\ \\
@ARTICLE$\{$lee2020pixelmpc,
\\author=$\{$Keuntaek $\{$Lee$\}$ and Jason $\{$Gibson$\}$ and Evangelos A. $\{$Theodorou$\}$$\}$,
\\journal=$\{$IEEE Robotics and Automation Letters$\}$,
\\title=$\{$$\{$Aggressive Perception-Aware Navigation using Deep Optical Flow Dynamics and PixelMPC$\}$$\}$,
\\year=$\{$2020$\}$
\\$\}$

\end{document}